\documentclass[letterpaper, 10 pt, conference]{ieeeconf} 
\IEEEoverridecommandlockouts                              
\usepackage{cite}
\usepackage{graphics}    
\usepackage{times}       
\usepackage{amsmath}     
\usepackage{amssymb}     
\usepackage{graphicx}
\usepackage{algorithm}
\usepackage{wrapfig}
\usepackage[noend]{algpseudocode}
\usepackage[font=small,labelfont=bf,skip = 2pt]{caption}
\usepackage[font=small,labelfont=bf,skip = 2pt]{subcaption}
\usepackage{comment}
\usepackage{hyperref}

\title{\LARGE \bf Handling Sparse Rewards in Reinforcement Learning \\ Using Model Predictive Control}
\author{Murad Dawood \and Nils Dengler \and  Jorge de Heuvel  \and Maren Bennewitz 
\thanks{All authors are with the Humanoid Robots Lab, University of Bonn,  Germany. Murad Dawood and Maren Bennewitz are additionally with the Lamarr Institute for Machine Learning and Artificial Intelligence, Germany.
This work has partially been funded by the Deutsche Forschungs- gemeinschaft (DFG, German Research Foundation) under the grant number BE 4420/2-2 (FOR 2535 Anticipating Human Behavior).}%
}

\def\figref#1{Fig.~\ref{#1}}
\def\tabref#1{Tab.~\ref{#1}}
\def\eqref#1{Eq.~(\ref{#1})}

\begin{document}
\maketitle
\thispagestyle{empty} 
\pagestyle{empty}

\begin{abstract} 
 Reinforcement learning (RL) has recently proven great success in various domains. Yet, the design of the reward function requires detailed domain expertise and tedious fine-tuning to ensure that agents are able to learn the desired behaviour. Using a sparse reward conveniently mitigates these challenges. However, the sparse reward represents a challenge on its own, often resulting in unsuccessful training of the agent. In this paper, we therefore address the sparse reward problem in RL. Our goal is to find an effective alternative to reward shaping, without using costly human demonstrations, that would also be applicable to a wide range of domains. Hence, we propose to use model predictive control~(MPC) as an experience source for training RL agents in sparse reward environments. Without the need for reward shaping, we successfully apply our approach in the field of mobile robot navigation both in simulation and real-world experiments with a Kuboki Turtlebot 2. We furthermore demonstrate great improvement over pure RL algorithms in terms of success rate as well as number of collisions and timeouts. 
Our experiments show that MPC as an experience source improves the agent's learning process for a given task in the case of sparse rewards.  
\end{abstract}

\section{Introduction}
\label{sec:intro}

Reinforcement learning~(RL) as well as model predictive control~(MPC) have been applied lately to various fields and shown impressive results. However, there are still great challenges that need to be dealt with in both approaches. One major challenge in RL is the design of the reward function. Shaping the reward function to achieve desired results requires lots of trials to get the expected behaviour of the trained policy. This is mainly due to the fact that during the training, the agents exploit any opportunity given by the reward function. An obvious solution to this issue would be to use sparse rewards, i.e., rewarding the agent only for achieving the goal and giving zero rewards otherwise. While this approach encourages the agent to complete a certain task, it is more difficult for the agent to identify promising behaviour. Since the agent has no idea how well it is performing during the training before reaching the goal, it may fail to find the optimal policy. Handling sparse rewards has been an active topic in the field of reinforcement learning~\cite{andrychowicz2017hindsight,vecerik2017leveraging, wang2020deep, agarwal2021goal, pfeiffer2018reinforced}. However, it still remains an open question how an RL agent be successfully trained in a sparse reward setting using an approach that is applicable to a variety of domains.

\begin{figure}[t]
\vspace{6pt}
  \centering
 \includegraphics[width=\linewidth]{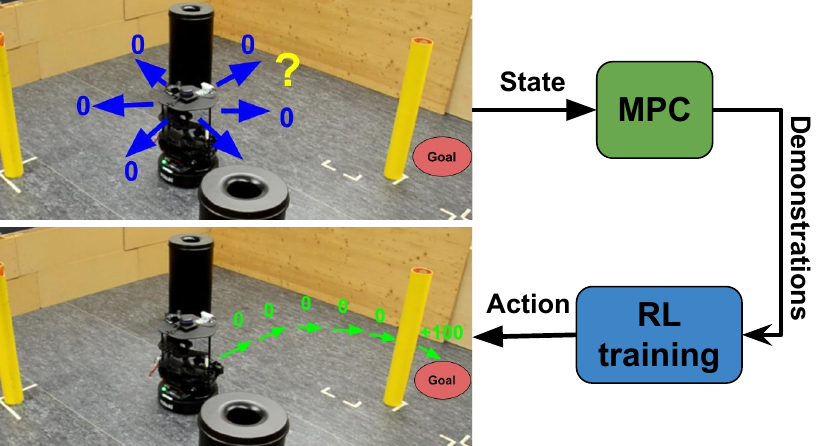}
 \caption{Handling sparse rewards using our approach. The model predictive controller~(MPC) provides demonstrations to the reinforcement learning~(RL) agent during training, while exploration still takes place. The MPC demonstrations in combination with RL guide the agent to find better policies to reach its goal.
 \vspace{-12pt}}  \label{fig:coverfig}
\end{figure}


One possibility is to use demonstrations to provide the agent with a course of actions that solve the task at hand. While demonstrations have been shown to improve the training process in case of sparse rewards \cite{yi2018deep, goecks2019integrating, hester2017learning}, and human demonstrations specifically are commonly used in the literature, providing these demonstrations can be quite costly. In addition to that, human demonstrations typically require hardware equipment or virtual reality sets to provide the demonstrations \cite{yi2018deep, delpreto2020helping, nair2018overcoming, DeHeuvel22roman}. In this work, we therefore propose to use MPC as an experience source for RL in the case of sparse rewards. MPC has been very popular lately in robotics and industry \cite{yao2018experimental, carlos2020efficient, osman2020end, dawood2020nonlinear, lucia2018deep, kloeser2020nmpc} as it is able to handle constraints on both states and control signals, can handle multiple-input multiple-output systems as well as nonlinear systems, and the cost function can be constructed in a straightforward way by minimizing the deviation between the reference states and the current states. The aim of our work is therefore to show that MPC can be used to provide demonstrations for an RL agent in sparse reward settings.

\begin{figure*}[t]
  \centering
 \includegraphics[width=0.8\linewidth]{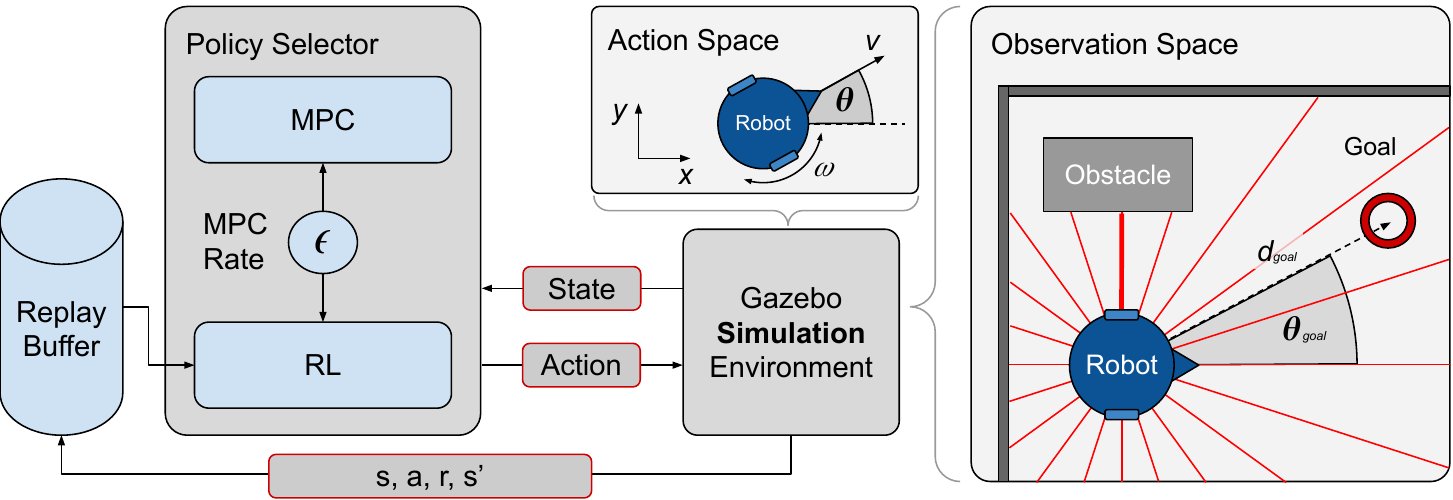}
 \caption{Illustration of our approach. \textbf{Left:} Both the model predictive control~(MPC) and reinforcement learning~(RL) interact with the simulated environment during training. Based on the MPC rate $\epsilon$, the policy selector chooses between MPC demonstrations and RL. The gathered tuples $(s, a, r, s')$ are stored in the replay buffer. \textbf{Top middle:} The robot's kinematics are used for nonlinear MPC. \textbf{Right:} The observation space includes a laser scan (red lines), the closest obstacle (thick red line), and the relative distance and heading to the goal. 
 \vspace{-10pt}}  \label{fig:architecture}
\end{figure*}

The motivation of using RL with MPC demonstrations is as follows: First, MPC is computationally demanding since it solves an optimization control problem at each time step. For highly nonlinear models with numerous states, this may not be feasible to run in real time on real-world applications~\cite{lucia2018deep}. In contrast, inferring a trained policy online for actions is less demanding even for systems with large state spaces. Second, while MPC can be tuned to satisfying performance in a certain scenario, the performance will not be as satisfying when the same controller is deployed in another scenario. This becomes obvious in trajectory tracking, where the weight matrices have to be further tuned for different trajectories~\cite{mehrez2013stabilizing}. Third, unlike MPC which demands the full state of the robot dynamic model, RL agents need only to attain partial observations from the environment which can be provided using onboard sensors~\cite{zhang2016learning}.

We investigate our approach in the field of robot navigation for the following reasons: 1) For MPC, the kinematic prediction model of mobile robots is straightforward. 2) The state and action space of mobile robots is small in comparison to, e.g., humanoids, so that tuning of the MPC is not time-consuming. 3) The learned policy can easily be tested in different scenarios and usually successfully be transferred to a real mobile robot. 

To summarize, our main contribution is to demonstrate that MPC as an experience source improves the training process of RL agents in sparse rewards settings. We showcase our approach in a mobile robot navigation scenario with static and dynamic obstacles, both in simulations and on a real robot. We also perform an ablation study to analyze the effect of varying the number of MPC demonstrations during the training. We make the following key claims: 
(i) MPC guides the RL agent to learn tasks in a pure sparse reward setting.
(ii) The learned behavior policy leads to higher success rates than pure RL.
(iii) The balance between MPC demonstrations and RL exploration influences the convergence rate of the training.
(iv) Our approach can successfully be applied to the task of mobile robot navigation.

\section{Related Work}
\label{sec:related}

We will first discuss the use of human demonstrations in the context of RL, followed by non-human demonstrations, and finally previous approaches combining MPC with RL.

\textbf{Human Demonstrations in Reinforcement Learning: }Several approaches have been presented that use human expert demonstrations to boost the training process of RL agents by showing examples of how to perform a certain task. The agents subsequently learn faster in comparison to exploring randomly. For example, \cite{hester2018deep} used human demonstrations to boost the training of deep Q-networks and showed great improvement over different RL approaches in Atari games. A similar approach \cite{vecerik2017leveraging} used human demonstrations to improve the training of deep deterministic policy gradient~(DDPG) in robotics tasks. Recently, \cite{yi2018deep} and \cite{liu2022improved} proposed combining supervised learning with RL and providing expert samples to play a video game and control a self-driving vehicle, respectively. Additionally, \cite{nair2018overcoming} also combined supervised learning with RL and human demonstrations to perform robot arm tasks and showed that using demonstrations outperforms the Hindsight Experience Replay~(HER) approach \cite{andrychowicz2017hindsight}.

\textbf{Non-Human Demonstrations: }To overcome the need for costly human demonstrations, several approaches for providing non-human demonstrations have been proposed. \cite{wang2020deep} applied a hand-crafted policy of low success rate to improve the training of an unmanned aerial vehicle~(UAV) in a sparse reward setting. As stated in that work, these hand-crafted policies cannot be applied to diverse scenarios since they are only able to perform fixed maneuvers. \cite{rengarajan2022reinforcement} used a partially trained RL agent with shaped rewards to provide demonstrations for another RL agent in sparse reward settings on MuJoCo \cite{todorov2012mujoco} simulations and to mobile robot navigation. \cite{xie2018learning}, \cite{wang2020learning} proposed using proportional controllers to provide the demonstrations for RL agents for mobile robot navigation and robotic arm manipulation, respectively. \cite{pfeiffer2018reinforced}~used demonstrations generated by a global planner to train a network using imitation learning along with RL for mobile robot navigation. Unlike the discussed approaches, we use a model predictive controller as an experience source since MPC can be applied to a variety of applications and does not involve reward shaping.

\textbf{Combining MPC with RL: }Several approaches using MPC and differential dynamic programming~(DDP) along with RL have been presented. In \cite{zhang2016learning} and \cite{levine2013guided} the authors implemented the guided policy search~(GPS) approach where they transform the RL problem into a supervised learning problem using demonstrations from MPC and DDP respectively to train a UAV and MuJoCo environments. \cite{bellegarda2020online} used MPC as an experience source and trained their network using supervised learning for the navigation of a simulated car model. However, their approach keeps the MPC running as a safe fail policy in case the RL agent fails to find a better action than the MPC. In \cite{shin2022infusing} the authors proposed to apply meta reinforcement learning along with MPC for demonstrations to train a mobile robot navigate through randomly moving obstacles. The authors used shaped rewards and the MPC is always running in case the agent cannot find an action specially when the robot is close to obstacles or the goal location. Furthermore, \cite{silver2018residual} trained the RL agent as a higher layer on top of the MPC to provide correction actions for the MPC to push objects using a robot arm. Unlike the previous approaches, we rely solely on RL training to learn from the MPC demonstrations and do not include supervised learning loss. Furthermore, we use a sparse reward setting to avoid reward shaping.

In \cite{turrisi2020enforcing} the authors applied DDPG after being trained offline to calculate a reference trajectory that is tracked by the MPC to control a Pendubot. \cite{lee2018safe} deployed imitation learning to learn from the MPC in cart-pole and autonomous driving scenarios. If the uncertainty of the network is high, the control is given solely to the MPC. To learn the threshold of uncertainty for switching the control the authors applied an RL agent. Furthermore, \cite{brito2021go} trained an RL agent to generate subgoals that are tracked using a MPC controller. This approach was implemented in a robot navigation scenario. Despite the improved performance of these approaches over pure RL methods, all of them require the MPC to be running the whole time which can be computationally demanding. In our work, we only use the MPC during the training process. During testing solely the learned RL policy is applied without the need for MPC as a fallback policy.

To the best of our knowledge, using MPC as an experience source to improve the training of RL agents in the case of sparse rewards has not been tackled before.
\section{Our Approach}
\label{sec:main}

The goal of our work is to run the RL agent independently of the MPC after training, while relying on the MPC demonstrations before. During training, we use a parallel architecture where the MPC and the RL agent run simultaneously and only one of the two output actions is chosen. 
To demonstrate our approach, we focus on solving mobile robot navigation around obstacles using only the laser scan as input to the RL agent. We run the chosen policy for the whole episode to provide demonstrations showing how the robot can navigate from the start position to the goal while avoiding obstacles. To choose between the MPC action and the RL action, we define the MPC rate $\epsilon$ that determines whether the MPC output or the RL action is taken for the whole upcoming episode. In the following, we present the components of our architecture in detail, a schematic overview can be found in \figref{fig:architecture}.

\begin{figure*}[ht]
  \centering
  \begin{subfigure}{0.31\textwidth} 
  \centering
    \includegraphics[width=\linewidth]{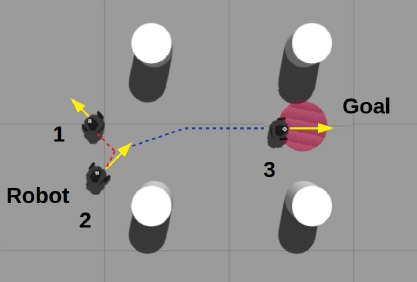}
  \subcaption{MPC\textunderscore SAC in static environment} \label{fig:sac_nav}
  \end{subfigure}
  \hfill
  \begin{subfigure}{0.65\textwidth} 
  \centering
 \includegraphics[width=\linewidth]{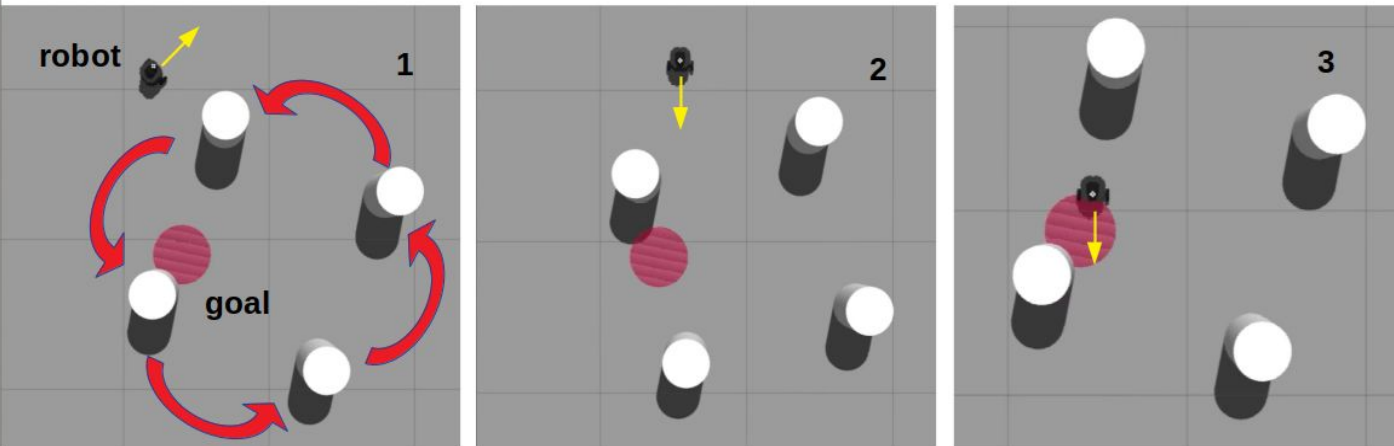}
 \subcaption{MPC\textunderscore SAC in dynamic environment}\label{fig:dyn_nav}
 \end{subfigure}
  \caption{Birds-eye view of the MPC\textunderscore SAC agent navigating from a random position to the goal (round red patch) in case of \textbf{(a)} static obstacles and \textbf{(b)} dynamic obstacles, both represented by white cylinders. The yellow arrows indicate the heading of the robot at different instances, the red trail indicates backward motion, while the blue trail indicates forward motion. The numbers denote the temporal order.
  \vspace{-10pt}}
\end{figure*}
\subsection{Nonlinear Model Predictive Control}

The nonlinear model predictive control~(NMPC) is a variant of the MPC that can handle nonlinear systems. In MPC schemes, there are two main components: the prediction model and the optimal control problem~(OCP). 
The prediction model is the robot dynamics model that is used to calculate the future states of the robot in a receding horizon fashion, i.e., the next future states of the robot are estimated at each new time step. The OCP consists of the cost function that has to be minimized while respecting the control and state constraints of the robot. To implement the NMPC, we used acados \cite{Verschueren2019}, a software package that provides the building blocks for describing and solving nonlinear optimal control problems.\

\textbf{Prediction Model: }The choice of the prediction model is critical for the design of the NMPC. A high fidelity model can accurately represent the nonlinearities of the actual robot but at the cost of high computational cost. While a simple mathematical model can lower the computational cost but at the cost of less accurate predictions. In this work, we represent the mobile robot using a nonlinear kinematic model (\figref{fig:architecture}). The state space representation of the model $\dot{X} = f(X,U)$ is as follows:
\begin{equation}
\begin{array}{@{}l}
\dot{x} = vcos(\theta)\\
\dot{y} = vsin(\theta)\\
\dot{\theta}  = \omega
\end{array}
\end{equation}
where $X$ represents the state of the mobile robot, i.e., its 2D position $(x, y)$ in space, and the heading of the robot $\theta$. $U$ represents the controls which are the linear and angular velocities, $v$ and $\omega$, respectively.\

\textbf{Optimal Control Problem: }The OCP solved at each time step is formulated by setting the cost function and the constraints on the controls and the states and is as follows:
\begin{subequations}
\begin{align}
& \underset{\substack{x_{k=1:N},\\ u_{k=1:N-1}}}{\text{min}}
& & J = \sum_{k=1}^N \lVert x_{\mathit{ref}} - x_{k} \rVert^2_{\mathit{Q}} + \lVert u_{k} \rVert^2_{\mathit{R}} + \lVert \dfrac{1}{e^{{\mathit{dist}}_{\mathit{k}}^2}}\lVert^2_\mathit{D} \\
& \text{subject to} & & x_{0|k} = x_{k} ,\\
& & & x_{i+1|k} = f(x_{i|k}, u_{i|k}) ,\\
& & &  u_{i|k} \in U ,\\
& & &  x_{i|k} \in X
\end{align}
\end{subequations}

\begin{table}
\centering
{\footnotesize
\begin{tabular}{|c|c|}
\hline
Parameter & Value  \\
\hline
$\mathit{Q}$ & diag([0.1, 0.1, 0.05])  \\
${\mathit{Q}}_{\mathit{e}}$ & diag([1, 1, 0.1])  \\
$\mathit{R}$ & diag([0.001, 0.01])  \\
$\mathit{D}$ & 20 \\
$\mathit{N}$ & 30\\
$\mathit{T}$ & 6 seconds\\
\hline
\end{tabular}
}
\caption{MPC parameters and weight matrices. ${\mathit{Q}}_{\mathit{e}}$ is the weight matrix at the final state $N$. $T$ is the prediction horizon in seconds. }\label{tab:params_mpc}
\vspace{-13pt}
\end{table}

Where $J$ represents the cost function to be minimized. The first term penalizes the difference between the predicted states ${\mathit{x}}_{\mathit{k}}$ and the target state ${\mathit{x}}_{\mathit{ref}}$ at instant $k$. The second term penalizes the control signals ${\mathit{u}}_{\mathit{k}}$. The final term penalizes the nearness to the closest obstacle, where ${\mathit{dist}}_{\mathit{k}}$ is the Euclidean distance to the nearest obstacle at instant~$k$. $Q$ and $R$ are diagonal weight matrices for the states and controls respectively. The weighting factor~$D$ balances the collision avoidance term. Constraint (2b) defines the initial state, (2c) forces the system's dynamic constraints, while (2d) and (2e) ensure that the controller satisfies the control and state limits at each time step. The weight matrices along with the MPC parameters are shown in \tabref{tab:params_mpc}. Note that the cost function uses the full state of the robot, while the RL uses only partial observations as defined in Section III C. Hence, the cost function is not usable as a reward function and the reward for the RL agent is still sparse.

\subsection{Reinforcement Learning Agent}
The foundation of reinforcement learning relies on the description of the world as a Markov decision process~(MDP), which is described by a tuple $M$: ~($S$, $A$, $R$, $P$, $\gamma$). Where $S$ is the set of states, $A$ is the set of actions, $R(s,a)$ is the reward function, $P(s'|s, a)$ is the trasition probability, and $\gamma$ is the discount factor. An agent in state $s \in S$ takes an action $a \in A$ resulting in the next state $s' \in S$, which is rewarded by reward $r$ and dsicounted by factor $\gamma$. The action $a$ is chosen according a policy $\pi$ that determines for each state which action the agent will take. The transition from state $s$ to state $s'$ upon taking action $a$ is determined by the transition probability $P$ which is also environment dependent. The main goal of the RL agent is then to maximize the total cumulative reward:
\begin{equation}
\begin{array}{@{}l}
$R$_{total} = \displaystyle\sum_{t=0}^{\infty} \gamma^{t} r_{t}
\end{array}
\end{equation}

We use an off-policy algorithm to control the mobile robot and use soft actor-critic (SAC) \cite{haarnoja2018soft} as the main agent. SAC maximizes the entropy of the policy along with the reward which resulted in a better exploration strategy and showed improved convergence rates compared to other algorithms in our experiments. 

\subsubsection{Soft Actor-Critic}
\begin{table}[tb]
\centering
{\footnotesize
\begin{tabular}{|c|c|}
\hline
Parameter & Value  \\
\hline
optimizer & Adam  \\
discount factor ($\gamma$) & 0.99  \\
replay buffer size & $10^6$ \\
hidden layers (all networks) & 2\\
hidden units per layer & 256\\
batch size & 256 \\
learning rate & $3.10^{-4}$\\
\hline
\end{tabular}
}
\caption{Parameters for SAC} \label{tab:params}
\vspace{-13pt}
\end{table}
The soft actor-critic aims to maximize the expected reward and the entropy of the policy by optimizing the following objective function:
\begin{equation}
\begin{aligned}
& J = \sum_{t=0}^TE_{(s_t,a_t)\sim\rho_{\pi}}[r(s_t,a_t) + \alpha H(\pi(.|s_t))], \\
\end{aligned}
\end{equation}
where $r$ is the reward the agent gets for executing action~$a_{t}$ at state $s_{t}$, $\pi$ is the policy, $\rho_{\pi}$ is the trajectory distribution induced by the policy $\pi$, $\alpha$ is the temperature of the entropy $H$, and $H(\pi(.|s_t)) = -E_{\pi}[\log \pi (.|s_t)]$ is the entropy of the policy.\

We used the same architecture as in~\cite{haarnoja2018soft} and also implemented the entropy regularization by optimizing $\alpha$ during the training. Our hyperparameters for the SAC are summarized in \tabref{tab:params}.

\subsubsection{MPC Rate}

The MPC rate \textbf{$\epsilon$} refers to the probability of choosing the MPC actions over the RL actions. Hence, this parameter can greatly affect the training performance since a high contribution of the MPC would mean less exploration of the RL agent and would result in a policy that acts more similarly to the MPC but is limited by the scenarios that only the MPC controller has experienced. A small~\textbf{$\epsilon$} would lead to more exploration and less demonstrations from the MPC. In our experiments, we provide an ablation study regarding different values for this parameter and its effect on the training.
We found it beneficial to decay the influence of the MPC over the course of training. Hence, we decay \textbf{$\epsilon$} over episodes by a decay rate of 0.5\% as the episodes~($n$) progress:
\begin{equation}
\begin{aligned}
& \epsilon_n = \epsilon_0 (0.995)^{n/4} \\
\end{aligned}
\label{eq:decay}
\end{equation}
Where $\epsilon_0$ is the initial value for the MPC rate at the beginning of the training and $\epsilon_n$ is the updated MPC rate.
\begin{figure*} [ht]
	\begin{subfigure}{0.6\linewidth} 
 		\includegraphics[width=\linewidth]{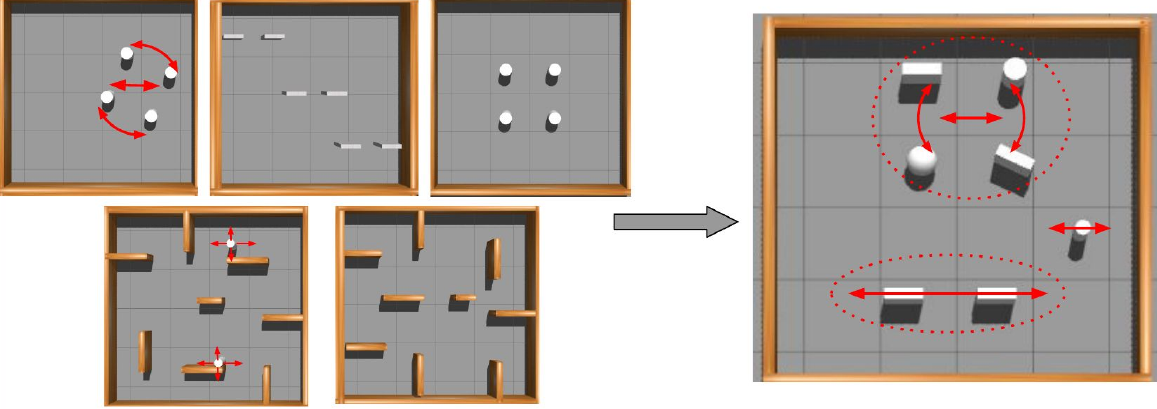}
 		\subcaption{Generalization test} \label{fig:Gener}
 	\end{subfigure}
 	\hfill
	 \begin{subfigure}{0.35\linewidth}
  	\centering
  		\includegraphics[width=\linewidth]{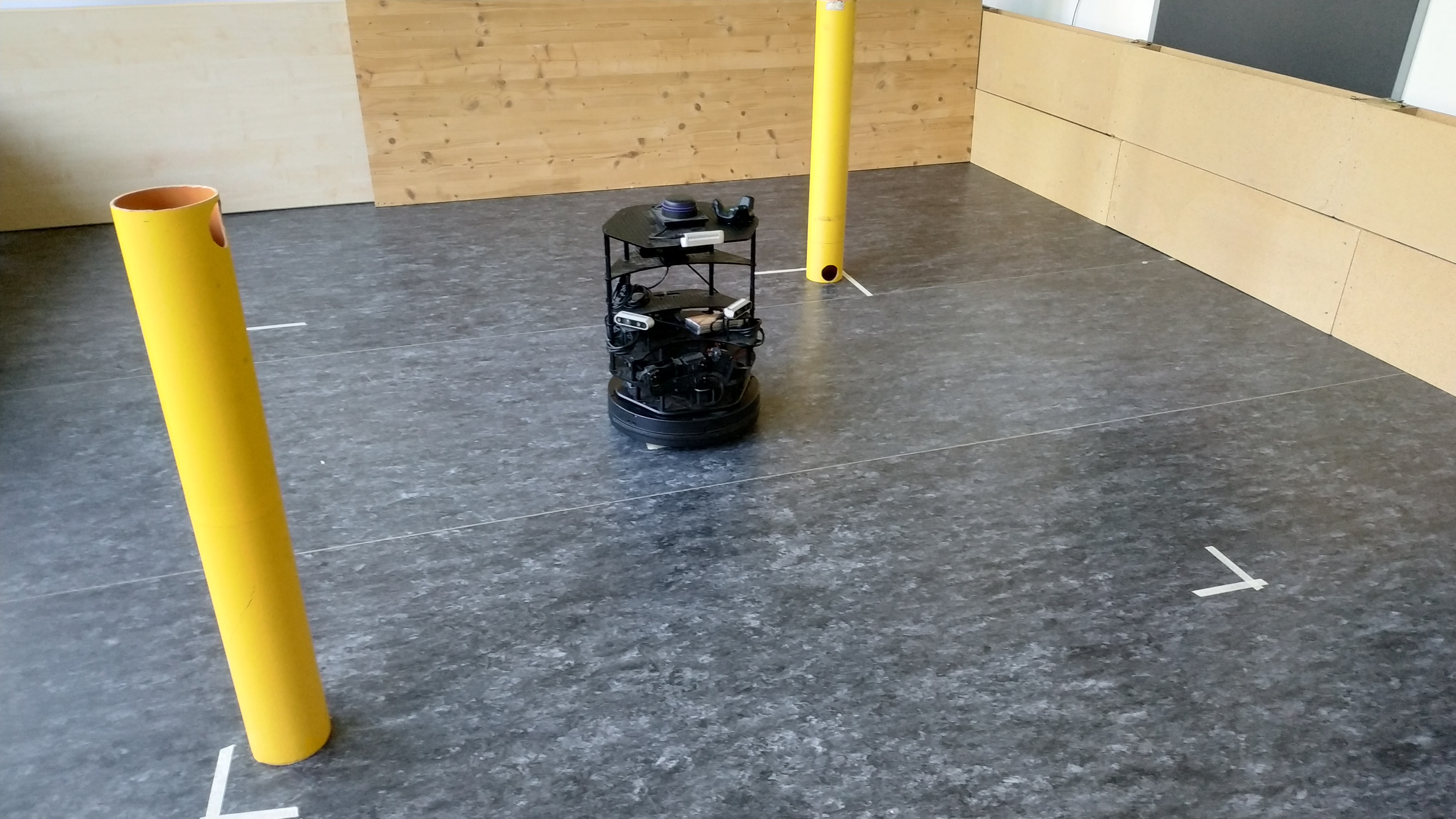}
 	 \subcaption{Real-world environment.}\label{fig:Real}
 	 \end{subfigure}
 \caption{\textbf{(a)} Generalization test: The robot is trained in the five left environments and is evaluated in the right one. The evaluation environment has moving walls and spheres which the agent has not jointly experienced during training. \textbf{(b)} The real environment for testing the Turtlebot has a size of $3.7\times 3.7\,m^2$ with two static obstacles. The goals are spawned randomly around the area. 
 \vspace{-13pt}}\label{fig:exps}
\end{figure*}
\subsection{Implementation of the RL Agent}
For the implementation of the RL agent in the navigation task, we defined the following action space, observation space, and reward function~(see also \figref{fig:architecture}).\\
\textbf{Action space:} We use continuous action spaces for both the linear and angular velocities of the mobile robot. The limits for the linear and angular velocities are $[-0.5\, m/s, 0.5\, m/s]$ and $[-0.785\, \mathit{rad}/s, 0.785\, \mathit{rad}/s]$, respectively.\\
\textbf{Observation space:} At each time step, the RL agent receices a scan of 20 lidar distance measurements, as well as the distance and heading difference to the target location as input. We furthermore provide the explicit distance and heading to the nearest obstacle as calculated from the laser scan.\\
\textbf{Rewards:} The sparse reward is defined as follows:
\begin{equation}
\begin{aligned}
& R = \begin{cases}
r_{\mathit{success}} &\text{if goal reached}, \\
r_{\mathit{collision}} &\text{if collision or stuck},\\
0 &\text{otherwise.}
\end{cases}
\end{aligned}
\end{equation}

\newpage
\section{Experimental Evaluation}
\label{sec:exp}

The main focus of this work is to show how MPC can improve the training of RL agents in a sparse reward setting. We use Gazebo \cite{koenig2004design} in combination with ROS~\cite{quigley2009ros} as a simulator during training and evaluation, and as a real-robot platform the Kuboki Turtlebot 2. We applied an MPC rate of $25\%$ during the experiments. To show that the MPC guides the RL agent to reach better behavior policies than pure RL in a sparse reward settings, and that our approach can be applied to mobile robot navigation, we evaluate our approach in the following scenarios:
(i)~A static environment, where the robot has to navigate around obstacles to reach different target positions. 
(ii)~A dynamic environment, where four dynamic obstacles are rotating in the environment and the robot has to navigate between them to reach the target.
(iii)~We test whether our agent is able to generalize to unseen scenarios, i.e., we trained the agent in different environments and evaluate its performance in a further, unseen environment.
(iv)~Finally, we evaluate the learned policy on a real robot. The experiments can be seen in the video\footnote{\url{https://youtu.be/Au6R92JHH5Q}}.

\subsection{Static Environment}

The first experiment shows the performance of our approach in the case of static obstacles. The environment is shown in \figref{fig:sac_nav}. We tested our approach with soft actor critic~(SAC) as described before and, additionally, with twin delayed deep deterministic policy gradient (TD3) \cite{fujimoto2018addressing} to show the general applicability. 
We trained the agents with and without the MPC for 7,000~episodes, where each episode ran for 1,000 steps. During each episode, we spawned a new random goal every time the previous goal had been reached. An episode terminates if the agent collides with an obstacle or if the robot is stuck in place for 20 consecutive steps. \\ As a further baseline, we implemented guided policy search~(GPS)~\cite{levine2013guided,zhang2016learning} and trained a deep neural network in a supervised learning manner. We use the partially observed state (same state used by the RL agents) as the input to the network to predict the actions and use the actions from the MPC to calculate the mean squared error loss for the network. Simultaneously, we minimize the deviation of the MPC action from the inferred action at each time step.\\ In \figref{fig:sac_mpc_static} we show the convergence curves for our approach with SAC (cf. MPC\_SAC) plotted against the pure SAC. As can be seen, our MPC\_SAC shows better performance than the pure SAC in terms of collected reward.

For evaluation of the trained agents, we spawned a fixed sequence of 70~random goals and calculated the success rate, collision rate, and timeout rate, the results are shown in \tabref{tab:success} and demonstrate that the MPC successfuly guides the RL agent to learn tasks in a sparse reward setting and that the learned policy has higher success rates than pure RL. 
The SAC on its own was able to reach the targets if it is spawned directly in front of the robot, but fails to navigate through the obstacles to reach further targets. Introducing the MPC to the training process makes the agent able to navigate around the obstacles to reach its targets.\
While the pure TD3 was stuck most of the time, with the MPC it was able to reach an impressive performance, i.e., the agent showed higher success rates as well as fewer collisions and timeouts. 
Also, GPS outperforms the pure RL agents, however, our approach shows superior performance compared to GPS, which can be credited to the exploration involved in the RL training.
\subsection{Dynamic Environment}

\begin{figure}
\centering
  \begin{subfigure} {0.8\linewidth} 
  	\includegraphics[width=\linewidth]{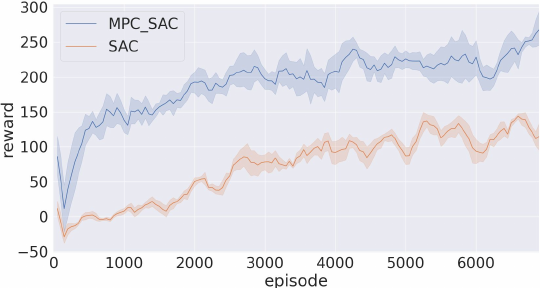}
  \subcaption{SAC vs MPC\textunderscore SAC in static environment (\figref{fig:sac_nav}).}   \label{fig:sac_mpc_static}
    \end{subfigure}
    \\[1.5ex]
  \begin{subfigure} {0.8\linewidth} 
  	\centering
  	\includegraphics[width=\linewidth]{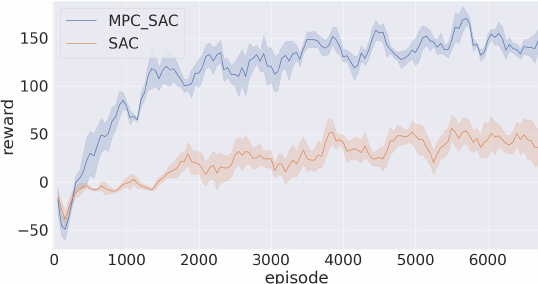}
  \subcaption{SAC vs MPC\textunderscore SAC in dynamic environment (\figref{fig:dyn_nav}).}   \label{fig:sac_mpc_dyn}
    \end{subfigure}
    \\[1.5ex]
  \begin{subfigure} {0.8\linewidth} 
  	\centering
  	\includegraphics[width=\linewidth]{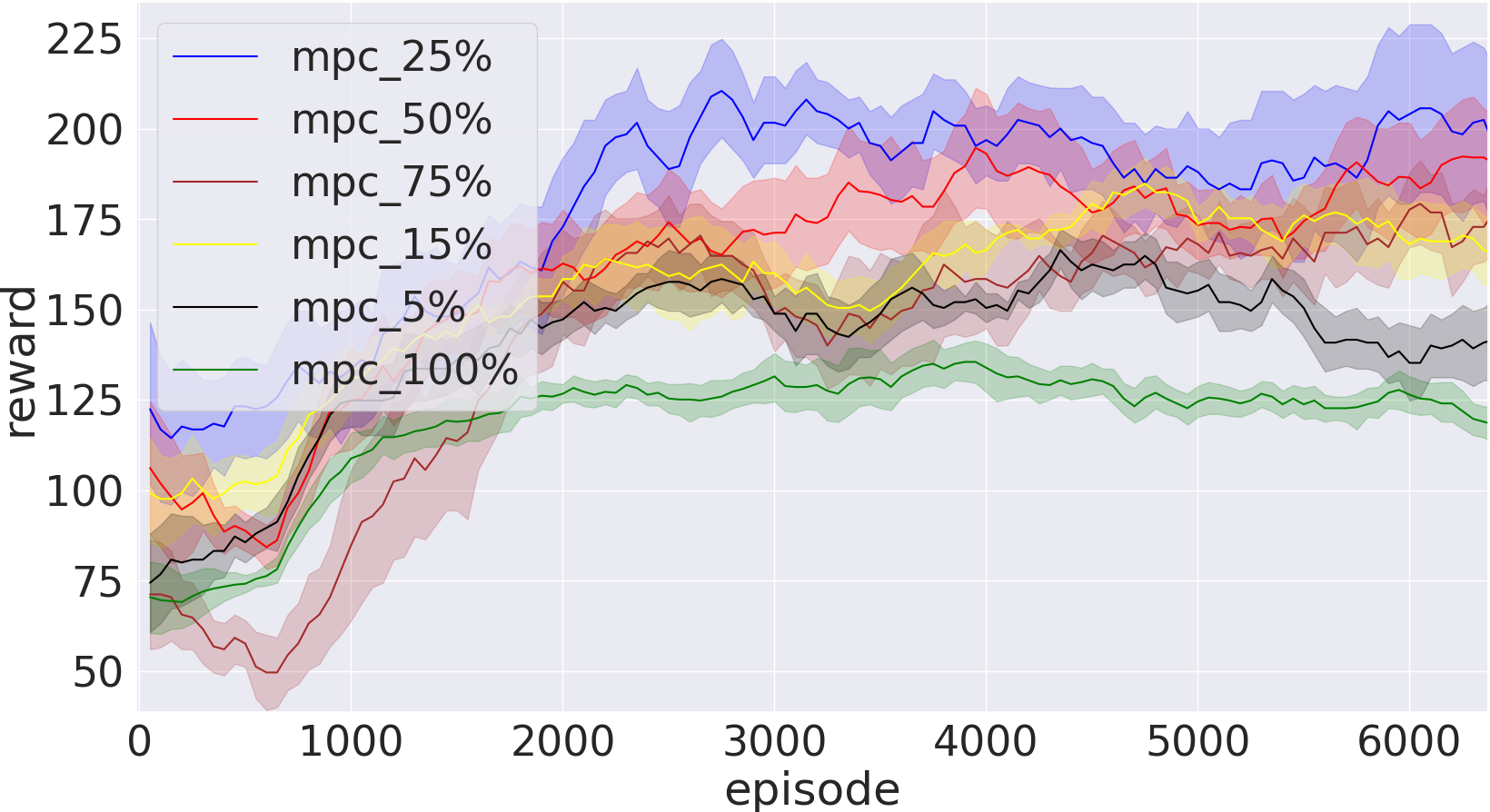}
    \subcaption{Ablation study for $\epsilon_0$ in static environment}  \label{fig:ablation}
      \end{subfigure}
      \\[1ex]
  \caption{Training performance for different setups. \textbf{(a), (b)} Including MPC as an experience source significantly increases the training performance by means of collected reward. \textbf{(c)} We found an initial MPC rate of 25\% to give the best results. The bold lines show the mean of five different seeds while the shaded areas as the standard error. }
 \label{fig:parameval}
\end{figure}

In the second experiment, we tested our approach in scenarios with dynamic obstacles, i.e., four obstacles rotating in the environment~(see \figref{fig:dyn_nav}) where the robot has no information about the behavior of the obstacles.
The training performance shown in \figref{fig:sac_mpc_dyn} indicates that including the MPC improves the navigation performance per episode by one goal (+100 reward) on average.\\ This improvement is also reflected in the success rate during testing, which increased by almost 50\% (see \tabref{tab:success}), indicating the improvement achieved by introducing the MPC as demonstrations during training.
\begin{table*}[t]
\centering
\footnotesize
\resizebox{0.75\textwidth}{!}{
\begin{tabular}{|p{3cm}||p{3cm}|p{3cm}|p{3cm}|}
\hline
\multicolumn{4}{|c|}{\textbf{Static Environment}} \\
\hline
\hfil Agent &\hfil Success rate &\hfil Collisions &\hfil Timeout \\
\hline
\hfil SAC & \hfil 52.5\% & \hfil 14.7\% & \hfil 32.8\% \\
\hfil MPC\textunderscore SAC & \hfil \textbf{97.2\%} & \hfil 0\% & \hfil 2.8\% \\
\hline
\hfil TD3 & \hfil 34.4\% & \hfil 2.8\% & \hfil 62.8\% \\
\hfil MPC\textunderscore TD3 & \hfil \textbf{98.6\% }& \hfil 1.4\% & \hfil 0\% \\
\hline
\hfil GPS & \hfil 88.5\% & \hfil 0\% & \hfil 11.5\% \\
\hline

\multicolumn{4}{|c|}{\textbf{Dynamic Environment}} \\
\hline
\hfil SAC & \hfil 42.8\% & \hfil 34.4\% & \hfil 22.8\% \\
\hfil MPC\textunderscore SAC & \hfil \textbf{91.4\% }& \hfil 7.2\% & \hfil 1.4\%\\
\hfil GPS & \hfil 82.8\% & \hfil 4.3\% & \hfil 12.9\% \\
\hline
\multicolumn{4}{|c|}{\textbf{Generalization to Different Environments}} \\
\hline
\hfil MPC\textunderscore SAC & \hfil 68.6\% & \hfil 26.7\% & \hfil 4.7\% \\
\hline
\multicolumn{4}{|c|}{\textbf{Real-Robot Experiments}} \\
\hline
\hfil MPC\textunderscore SAC &\hfil 85 $\pm$ 5\% & \hfil 6.7 $\pm$ 2.3\% & \hfil 8.3 $\pm$ 2.3\% \\
\hline
\end{tabular} }
\caption{Evaluation in different test scenarios. For each scenario, a fixed sequence of random goals was spawned. Timeout is the fraction of goals that the agent missed to reach within a fixed number of time steps. Our approach using the MPC experience outperforms both pure RL (SAC\cite{haarnoja2018soft} and TD3\cite{fujimoto2018addressing}) and GPS (Guided Policy Search)\cite{levine2013guided} approaches.
\vspace{-13pt}} \label{tab:success}
\end{table*}
\subsection{Generalization to Different Environments}

The aim of this experiment is to test whether the demonstrations provided by the MPC can aid the RL agent to handle unforeseen scenarios. We trained the agents in five different environments to gather diverse experience and then evaluate its performance in a sixth environment~(see~\figref{fig:Gener}). The results are also included in \tabref{tab:success}. The agent was unable to handle obstacles not experienced before such as moving walls and spheres, however, we found that the robot learned a primitive strategy that enables it to reach its target at some instances. More specifically, the agent waits for the moving obstacles until they do not block the way towards the goal anymore.

\subsection{Real-Robot Experiment}
  
Finally, we demonstrate that our approach is transferable to real mobile robots. We trained an agent in the static environment. To improve the sim-to-real transfer in terms of sensor noise, we added Gaussian noise ($\mu=0$, $\sigma=0.15$) to the lidar observation and Gaussian noise to the relative distance to the goal during training. The environment for the real-robot experiments is shown in \figref{fig:Real} and the mean success rate for three trials with 20~random goals each is included in \tabref{tab:success}. The robot had a mean success rate of 85\%, a mean collision rate of 6.7\% and a mean timeout rate of 8.3\%, over 3 runs.
\subsection{Ablation Study}

The balance between RL and MPC influences the training performance. Hence, we conducted an ablation study to evaluate different MPC rates $\epsilon_0$ for the contribution of the MPC controller. We investigate how including more or less MPC episodes affects the training of the RL agent (see \figref{fig:ablation}), i.e., we evaluated the performance for $\epsilon_0=$ [5\%, 15\%, 25\%, 50\%, 75\%, 100\%]. Note that those are the initial rates, which decay during training, see \eqref{eq:decay}. We experimentally found that a MPC rate of 25\% gave the best results in terms of reached rewards. This is a similar to \cite{shin2022infusing}, where an activation rate for the MPC of 20\% was found to be best. 

\section{Conclusion}
\label{sec:conclusion}
In this paper, we presented a novel approach to handle the challenges of sparse rewards in reinforcement learning~(RL) using model predictive control~(MPC) as an experience source. We show that the MPC demonstrations guide the RL agent to converge faster and find better policies. In the domain of robot navigation, our approach outperforms pure reinforcement learning algorithms in terms of success rate as well as number of collisions and timeouts. Furthermore, our ablation study shows the effect of varying the MPC rate on the training result. Finally, we showed that the learned controller can be successfully applied to a real robot.\\ 




\bibliographystyle{IEEEtran}
\bibliography{bibliography}

\begin{thebibliography}{10}
\providecommand{\url}[1]{#1}
\csname url@rmstyle\endcsname
\providecommand{\newblock}{\relax}
\providecommand{\bibinfo}[2]{#2}
\providecommand\BIBentrySTDinterwordspacing{\spaceskip=0pt\relax}
\providecommand\BIBentryALTinterwordstretchfactor{4}
\providecommand\BIBentryALTinterwordspacing{\spaceskip=\fontdimen2\font plus
\BIBentryALTinterwordstretchfactor\fontdimen3\font minus
  \fontdimen4\font\relax}
\providecommand\BIBforeignlanguage[2]{{%
\expandafter\ifx\csname l@#1\endcsname\relax
\typeout{** WARNING: IEEEtran.bst: No hyphenation pattern has been}%
\typeout{** loaded for the language `#1'. Using the pattern for}%
\typeout{** the default language instead.}%
\else
\language=\csname l@#1\endcsname
\fi
#2}}

\bibitem{andrychowicz2017hindsight}
M.~Andrychowicz, F.~Wolski, A.~Ray, J.~Schneider, R.~Fong, P.~Welinder,
  B.~McGrew, J.~Tobin, O.~Pieter~Abbeel, and W.~Zaremba, ``Hindsight experience
  replay,'' \emph{Advances in neural information processing systems}, 2017.

\bibitem{vecerik2017leveraging}
M.~Vecerik, T.~Hester, J.~Scholz, F.~Wang, O.~Pietquin, B.~Piot, N.~Heess,
  T.~Roth{\"o}rl, T.~Lampe, and M.~Riedmiller, ``Leveraging demonstrations for
  deep reinforcement learning on robotics problems with sparse rewards,''
  \emph{arXiv preprint arXiv:1707.08817}, 2017.

\bibitem{wang2020deep}
C.~Wang, J.~Wang, J.~Wang, and X.~Zhang, ``Deep-reinforcement-learning-based
  autonomous uav navigation with sparse rewards,'' \emph{IEEE Internet of
  Things Journal}, 2020.

\bibitem{agarwal2021goal}
P.~Agarwal, P.~de~Beaucorps, and R.~de~Charette, ``Goal-constrained sparse
  reinforcement learning for end-to-end driving,'' \emph{arXiv preprint
  arXiv:2103.09189}, 2021.

\bibitem{pfeiffer2018reinforced}
M.~Pfeiffer, S.~Shukla, M.~Turchetta, C.~Cadena, A.~Krause, R.~Siegwart, and
  J.~Nieto, ``Reinforced imitation: Sample efficient deep reinforcement
  learning for mapless navigation by leveraging prior demonstrations,''
  \emph{IEEE Robotics and Automation Letters (RA-L)}, 2018.

\bibitem{yi2018deep}
M.~Yi, X.~Xu, Y.~Zeng, and S.~Jung, ``Deep imitation reinforcement learning
  with expert demonstration data,'' \emph{The Journal of Engineering}, 2018.

\bibitem{goecks2019integrating}
V.~G. Goecks, G.~M. Gremillion, V.~J. Lawhern, J.~Valasek, and N.~R. Waytowich,
  ``Integrating behavior cloning and reinforcement learning for improved
  performance in dense and sparse reward environments,'' \emph{arXiv preprint
  arXiv:1910.04281}, 2019.

\bibitem{hester2017learning}
T.~Hester, M.~Vecerik, O.~Pietquin, M.~Lanctot, T.~Schaul, B.~Piot,
  A.~Sendonaris, G.~Dulac-Arnold, I.~Osband, J.~Agapiou, \emph{et~al.},
  ``Learning from demonstrations for real world reinforcement learning,''
  \emph{arXiv preprint arXiv:1704.03732}, 2017.

\bibitem{delpreto2020helping}
J.~DelPreto, J.~I. Lipton, L.~Sanneman, A.~J. Fay, C.~Fourie, C.~Choi, and
  D.~Rus, ``Helping robots learn: a human-robot master-apprentice model using
  demonstrations via virtual reality teleoperation,'' in \emph{Proc.~of the
  IEEE Intl.~Conf.~on Robotics \& Automation (ICRA)}.\hskip 1em plus 0.5em
  minus 0.4em\relax IEEE, 2020.

\bibitem{nair2018overcoming}
A.~Nair, B.~McGrew, M.~Andrychowicz, W.~Zaremba, and P.~Abbeel, ``Overcoming
  exploration in reinforcement learning with demonstrations,'' in
  \emph{Proc.~of the IEEE Intl.~Conf.~on Robotics \& Automation (ICRA)}.\hskip
  1em plus 0.5em minus 0.4em\relax IEEE, 2018.

\bibitem{DeHeuvel22roman}
J.~de~Heuvel, N.~Corral, L.~Bruckschen, and M.~Bennewitz, ``Learning
  personalized human-aware robot navigation using virtual reality
  demonstrations from a user study,'' in \emph{Proc.~of the IEEE Int.~Conf.~on
  Robot \& Human Interactive Communication (RO-MAN)}, 2022.

\bibitem{yao2018experimental}
F.~Yao, C.~Yang, X.~Liu, and M.~Zhang, ``Experimental evaluation on depth
  control using improved model predictive control for autonomous underwater
  vehicle (auvs),'' \emph{Sensors}, 2018.

\bibitem{carlos2020efficient}
B.~B. Carlos, T.~Sartor, A.~Zanelli, G.~Frison, W.~Burgard, M.~Diehl, and
  G.~Oriolo, ``An efficient real-time nmpc for quadrotor position control under
  communication time-delay,'' in \emph{Proc.~of the Int.~Conf.~on Control,
  Automation, Robotics and Vision (ICARCV)}.\hskip 1em plus 0.5em minus
  0.4em\relax IEEE, 2020.

\bibitem{osman2020end}
M.~Osman, M.~W. Mehrez, S.~Yang, S.~Jeon, and W.~Melek, ``End-effector
  stabilization of a 10-dof mobile manipulator using nonlinear model predictive
  control,'' \emph{IFAC-PapersOnLine}, 2020.

\bibitem{dawood2020nonlinear}
M.~Dawood, M.~Abdelaziz, M.~Ghoneima, and S.~Hammad, ``A nonlinear model
  predictive controller for autonomous driving,'' in \emph{Proc.~of the
  Intl.~Conf.~on Innovative Trends in Communication and Computer Engineering
  (ITCE)}.\hskip 1em plus 0.5em minus 0.4em\relax IEEE, 2020.

\bibitem{lucia2018deep}
S.~Lucia and B.~Karg, ``A deep learning-based approach to robust nonlinear
  model predictive control,'' \emph{IFAC-PapersOnLine}, 2018.

\bibitem{kloeser2020nmpc}
D.~Kloeser, T.~Schoels, T.~Sartor, A.~Zanelli, G.~Prison, and M.~Diehl, ``Nmpc
  for racing using a singularity-free path-parametric model with obstacle
  avoidance,'' \emph{IFAC-PapersOnLine}, 2020.

\bibitem{mehrez2013stabilizing}
M.~W. Mehrez, G.~K. Mann, and R.~G. Gosine, ``Stabilizing {NMPC} of wheeled
  mobile robots using open-source real-time software,'' in \emph{Proc.~of the
  Int.~Conf.~on Advanced Robotics (ICAR)}.\hskip 1em plus 0.5em minus
  0.4em\relax IEEE, 2013.

\bibitem{zhang2016learning}
T.~Zhang, G.~Kahn, S.~Levine, and P.~Abbeel, ``Learning deep control policies
  for autonomous aerial vehicles with mpc-guided policy search,'' in
  \emph{Proc.~of the IEEE Intl.~Conf.~on Robotics \& Automation (ICRA)}.\hskip
  1em plus 0.5em minus 0.4em\relax IEEE, 2016.

\bibitem{hester2018deep}
T.~Hester, M.~Vecerik, O.~Pietquin, M.~Lanctot, T.~Schaul, B.~Piot, D.~Horgan,
  J.~Quan, A.~Sendonaris, I.~Osband, \emph{et~al.}, ``Deep {Q}-learning from
  demonstrations,'' in \emph{Proc.~of the Conference on Advancements of
  Artificial Intelligence (AAAI)}, 2018.

\bibitem{liu2022improved}
H.~Liu, Z.~Huang, J.~Wu, and C.~Lv, ``Improved deep reinforcement learning with
  expert demonstrations for urban autonomous driving,'' in \emph{2022 IEEE
  Intelligent Vehicles Symposium (IV)}.\hskip 1em plus 0.5em minus 0.4em\relax
  IEEE, 2022.

\bibitem{rengarajan2022reinforcement}
D.~Rengarajan, G.~Vaidya, A.~Sarvesh, D.~Kalathil, and S.~Shakkottai,
  ``Reinforcement learning with sparse rewards using guidance from offline
  demonstration,'' \emph{arXiv preprint arXiv:2202.04628}, 2022.

\bibitem{todorov2012mujoco}
E.~Todorov, T.~Erez, and Y.~Tassa, ``Mujoco: A physics engine for model-based
  control,'' in \emph{Proc.~of the IEEE/RSJ Intl.~Conf.~on Intelligent Robots
  and Systems (IROS)}.\hskip 1em plus 0.5em minus 0.4em\relax IEEE, 2012.

\bibitem{xie2018learning}
L.~Xie, S.~Wang, S.~Rosa, A.~Markham, and N.~Trigoni, ``Learning with training
  wheels: speeding up training with a simple controller for deep reinforcement
  learning,'' in \emph{Proc.~of the IEEE Intl.~Conf.~on Robotics \& Automation
  (ICRA)}.\hskip 1em plus 0.5em minus 0.4em\relax IEEE, 2018.

\bibitem{wang2020learning}
G.~Wang, M.~Xin, W.~Wu, Z.~Liu, and H.~Wang, ``Learning of long-horizon
  sparse-reward robotic manipulator tasks with base controllers,'' \emph{arXiv
  e-prints}, 2020.

\bibitem{levine2013guided}
S.~Levine and V.~Koltun, ``Guided policy search,'' in \emph{International
  conference on machine learning}.\hskip 1em plus 0.5em minus 0.4em\relax PMLR,
  2013, pp. 1--9.

\bibitem{bellegarda2020online}
G.~Bellegarda and K.~Byl, ``An online training method for augmenting mpc with
  deep reinforcement learning,'' in \emph{Proc.~of the IEEE/RSJ Intl.~Conf.~on
  Intelligent Robots and Systems (IROS)}.\hskip 1em plus 0.5em minus
  0.4em\relax IEEE, 2020.

\bibitem{shin2022infusing}
J.~Shin, A.~Hakobyan, M.~Park, Y.~Kim, G.~Kim, and I.~Yang, ``Infusing model
  predictive control into meta-reinforcement learning for mobile robots in
  dynamic environments,'' \emph{IEEE Robotics and Automation Letters (RA-L)},
  2022.

\bibitem{silver2018residual}
T.~Silver, K.~Allen, J.~Tenenbaum, and L.~Kaelbling, ``Residual policy
  learning,'' \emph{arXiv preprint arXiv:1812.06298}, 2018.

\bibitem{turrisi2020enforcing}
G.~Turrisi, B.~B. Carlos, M.~Cefalo, V.~Modugno, L.~Lanari, and G.~Oriolo,
  ``Enforcing constraints over learned policies via nonlinear mpc: Application
  to the pendubot,'' \emph{IFAC-PapersOnLine}, 2020.

\bibitem{lee2018safe}
K.~Lee, K.~Saigol, and E.~A. Theodorou, ``Safe end-to-end imitation learning
  for model predictive control,'' \emph{arXiv preprint arXiv:1803.10231}, 2018.

\bibitem{brito2021go}
B.~Brito, M.~Everett, J.~P. How, and J.~Alonso-Mora, ``Where to go next:
  learning a subgoal recommendation policy for navigation in dynamic
  environments,'' \emph{IEEE Robotics and Automation Letters (RA-L)}, 2021.

\bibitem{Verschueren2019}
R.~Verschueren, G.~Frison, D.~Kouzoupis, J.~Frey, N.~van Duijkeren, A.~Zanelli,
  B.~Novoselnik, T.~Albin, R.~Quirynen, and M.~Diehl, ``acados: a modular
  open-source framework for fast embedded optimal control,'' 2020.

\bibitem{haarnoja2018soft}
T.~Haarnoja, A.~Zhou, K.~Hartikainen, G.~Tucker, S.~Ha, J.~Tan, V.~Kumar,
  H.~Zhu, A.~Gupta, P.~Abbeel, \emph{et~al.}, ``Soft actor-critic algorithms
  and applications,'' \emph{arXiv preprint arXiv:1812.05905}, 2018.

\bibitem{koenig2004design}
N.~Koenig and A.~Howard, ``Design and use paradigms for gazebo, an open-source
  multi-robot simulator,'' in \emph{Proc.~of the IEEE/RSJ Intl.~Conf.~on
  Intelligent Robots and Systems (IROS)}.\hskip 1em plus 0.5em minus
  0.4em\relax IEEE, 2004.

\bibitem{quigley2009ros}
M.~Quigley, K.~Conley, B.~Gerkey, J.~Faust, T.~Foote, J.~Leibs, R.~Wheeler,
  A.~Y. Ng, \emph{et~al.}, ``{ROS}: an open-source robot operating system,'' in
  \emph{ICRA workshop on open source software}, vol.~3.\hskip 1em plus 0.5em
  minus 0.4em\relax Kobe, Japan, 2009, p.~5.

\bibitem{fujimoto2018addressing}
S.~Fujimoto, H.~Hoof, and D.~Meger, ``Addressing function approximation error
  in actor-critic methods,'' in \emph{Proc.~of the Intl.~Conf.~on Machine
  Learning(ICML)}.\hskip 1em plus 0.5em minus 0.4em\relax PMLR, 2018.

\end{thebibliography}

\end{document}